\documentclass[letterpaper, 10 pt, conference]{ieeeconf}  

\IEEEoverridecommandlockouts                              

\usepackage{graphics} 
\usepackage{epsfig} 
\usepackage{subfigure}
\usepackage{enumerate}
\usepackage{amsmath} 

\title{\LARGE \bf
CPG-Based Control Scheme for Quadruped Robot to Withstand the Lateral Impact
}

\author{Qingsheng Luo$^{1}$, Chenyang Zhou$^{2}$, Yan Jia$^{1}$, Jianfeng Gao$^{1}$, Fangzheng Liu$^{3}$
\thanks{*This Paper is the extended English version of the Chinese journal paper \textbf{CPG-Based Control Scheme for Quadruped Robot to Withstand the Lateral Impact}, which firstly appeared in \emph{Transactions of Beijing Institute of Technology} 35.4 (2015): 384-390.}
\thanks{$^{1}$Qingsheng Luo, Yan Jia and Jianfeng Gao is with School of Mechatronical Engineering, Beijing Institute of Technology,  100081, Beijing, China
        {\tt\small }}%
\thanks{$^{2}$Chenyang Zhou is with School of Mechanical Engineering, Beijing Institute of Technology,  100081, Beijing, China
        {\tt\small }}%
\thanks{$^{3}$Fangzheng Liu is with School of Information and Electronics, Beijing Institute of Technology,  100081, Beijing, China
        {\tt\small }}%
}

\begin{document}

\maketitle
\thispagestyle{empty}
\pagestyle{empty}

\begin{abstract}

This paper aims to present a stability control strategy for quadruped robot under lateral impact with the help of lateral trot. We firstly propose five necessary conditions for keeping balance. The classical four-neuron Central Pattern Generator (CPG) network with Hopf oscillators is then extended to eight-neuron network with four more trigger-enabled neurons, which controls the lateral trot. With proper adjustment of network's parameters, such network can coordinate the lateral and longitudinal trot gait. Based on Zero Movement Point (ZMP) theory, the robot is modeled as an inverted pendulum to plan the Center of Gravity (CoG) position and calculate the needed lateral step length. The simulation shows that the lateral acceleration of the quadruped robot after lateral impact regains to the normal range in a short time. Comparison shows that the maximal lateral impact that robot can resist increases about 125\% from 0.72g to 1.55g.

\end{abstract}

\section{INTRODUCTION}

Bionic concept has penetrated into the robot mechanical design as well as control method. Many control methods mainly followed in the modeling-plan-control pattern, such as P-series of Honda, ASIMO\cite{c1},SDR-XX of Sony, ORIO\cite{c2} and so on, which required a lot of calculations. But a biologically inspired approach, namely central pattern generator (CPG) can significantly decrease the calculations. Kimura began to apply CPG to control robots rhythmic movement and introduced several reflex into the system in 1994\cite{c3}\cite{c4}. Gentaro Taga\cite{c5}\cite{c6} and Nagashima\cite{c7}as well as Tsuchiya K. used CPG to control dynamic walking of biped robots. Kimura and Fukuoka investigated how a quadruped robot walks in a unknown terrain\cite{c8}.However, the majority focuses on the traditional environment and the aspects such as gait strategy in relatively stable surroundings. Our purpose is to help our robot to withstand an unknown lateral impact. A well-known representative example is the Big Dog by Boston Dynamics, which has hydraulic cylinder driving joints to withstand lateral impact through laterally stepping\cite{c9}.

\begin{figure}[thpb]
      \centering
            \includegraphics[width = 2.5in]{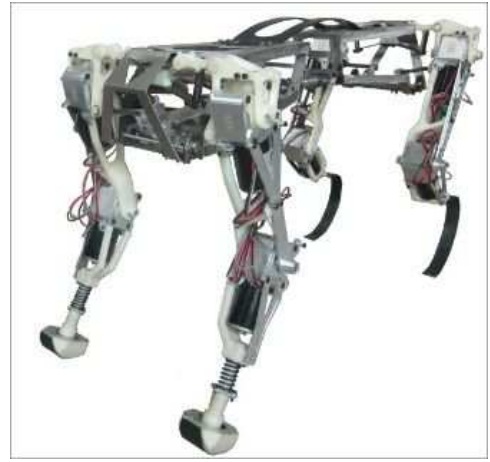}
      \caption{The Runner Robot}
      \label{fig:1}
   \end{figure}

We developed a quadruped robot called Runner, in Fig.\ref{fig:1}. Its hips combine the traditional forward moveable joints as well as the laterally moveable joints (LM joints), which are driven by industrial servos and DC motors respectively. Lateral trot is introduced to enable the robot to withstand the lateral impact. The classical CPG network is extended to coordinate the lateral and longitudinal trot. Based on the ZMP and inverted pendulum model, the step length of lateral trot can be calculated. This paper is organized as following: Section II introduces the mechanical system and the design concept. Section III and IV outline the whole system. The system is validated in section V. Conclusion and future work are presented in section VI.
\section{Concept}
\subsection{Mechanical Design of Laterally Movable Joints}
\begin{figure}[thpb]
      \centering
            \includegraphics[width = 3in]{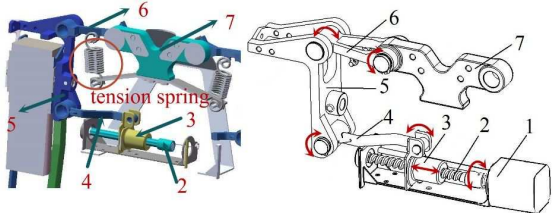}
      \caption{Laterally movable systems}
      \label{fig:2}
   \end{figure}

The laterally movable joints in Fig.\ref{fig:2} can be regarded as a five linkage (Part3, 4, 5, 6, 7). Part 1,2,7 are fixed on the body. Part 1 is a DC motor. The screw 2 transforms the rotational motion of the motor 1 into the linear motion of the sliding table 3. Parts 3,4,5 are connected by revolution joint. Part 6 is restrained by a tension spring. Part 5 is fixed with one leg. So in this mechanism, Part 3 is the only active part. If the tension spring are stiff enough (this means that part 6 and 7 are fixed together), then there is only one DOF left. With the linear movement of the sliding table, the leg can complete the lateral movement. 
The relationship between the foot's lateral displacement \(foot\) and linear sliding position \({x_{\rm{L}}}\) is approximated by:
\begin{equation}
foot =
\begin{cases}
5.5{x_L},  fore\_leg \\
6.08{x_L},  hind\_leg
\end{cases}
\label{eq:footLat}
\end{equation}

\subsection{Considerations for the Lateral Stability Control Strategy}
\begin{figure}[thpb]
    \begin{tabular}{cc}
    \begin{minipage}[t]{1.5in}
    \includegraphics[width = 1.5in]{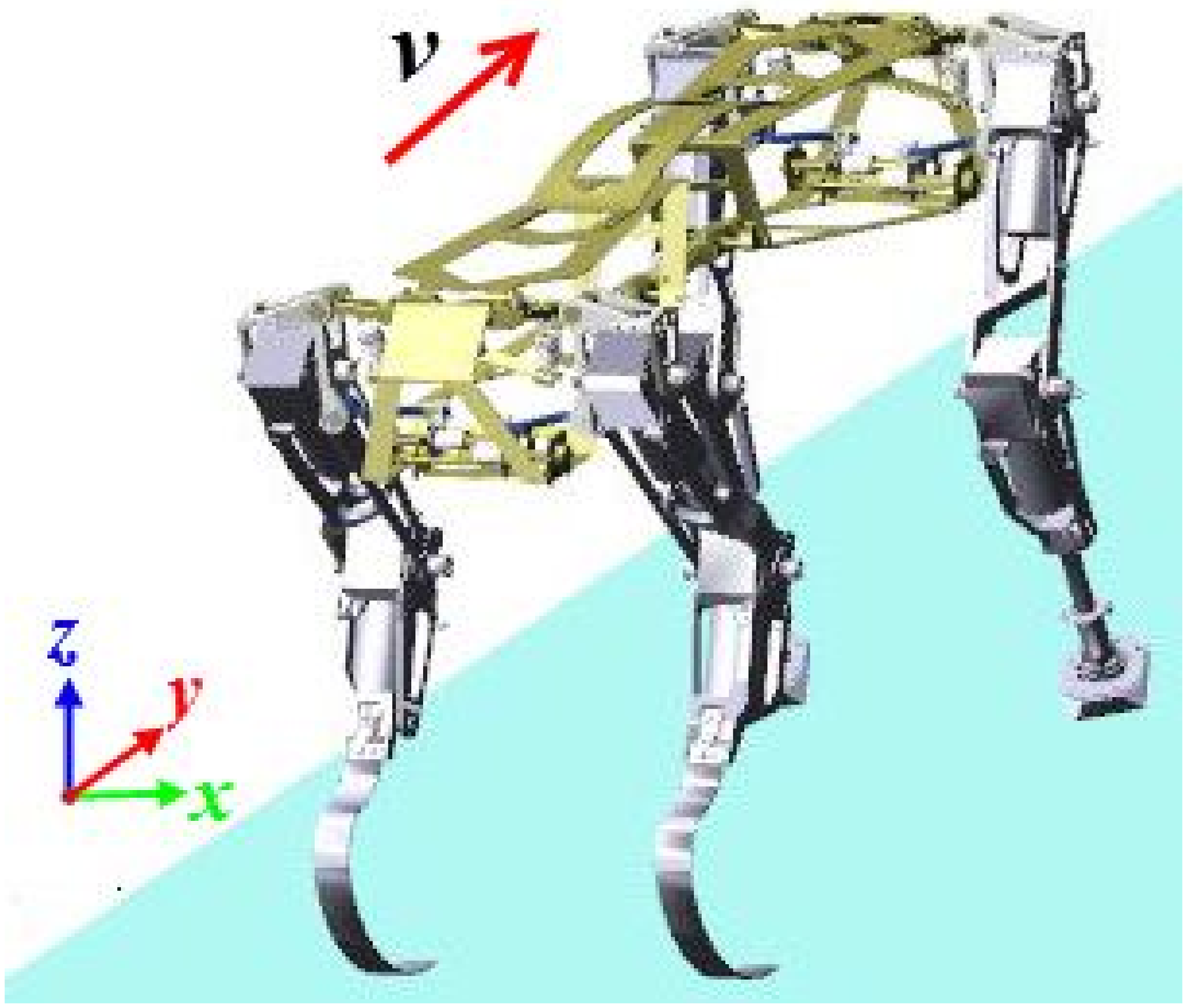}
    \caption{Runner Robot WCS}
    \label{fig:3}
    \end{minipage}
    \begin{minipage}[t]{1.5in}
   \includegraphics[width = 1.5in]{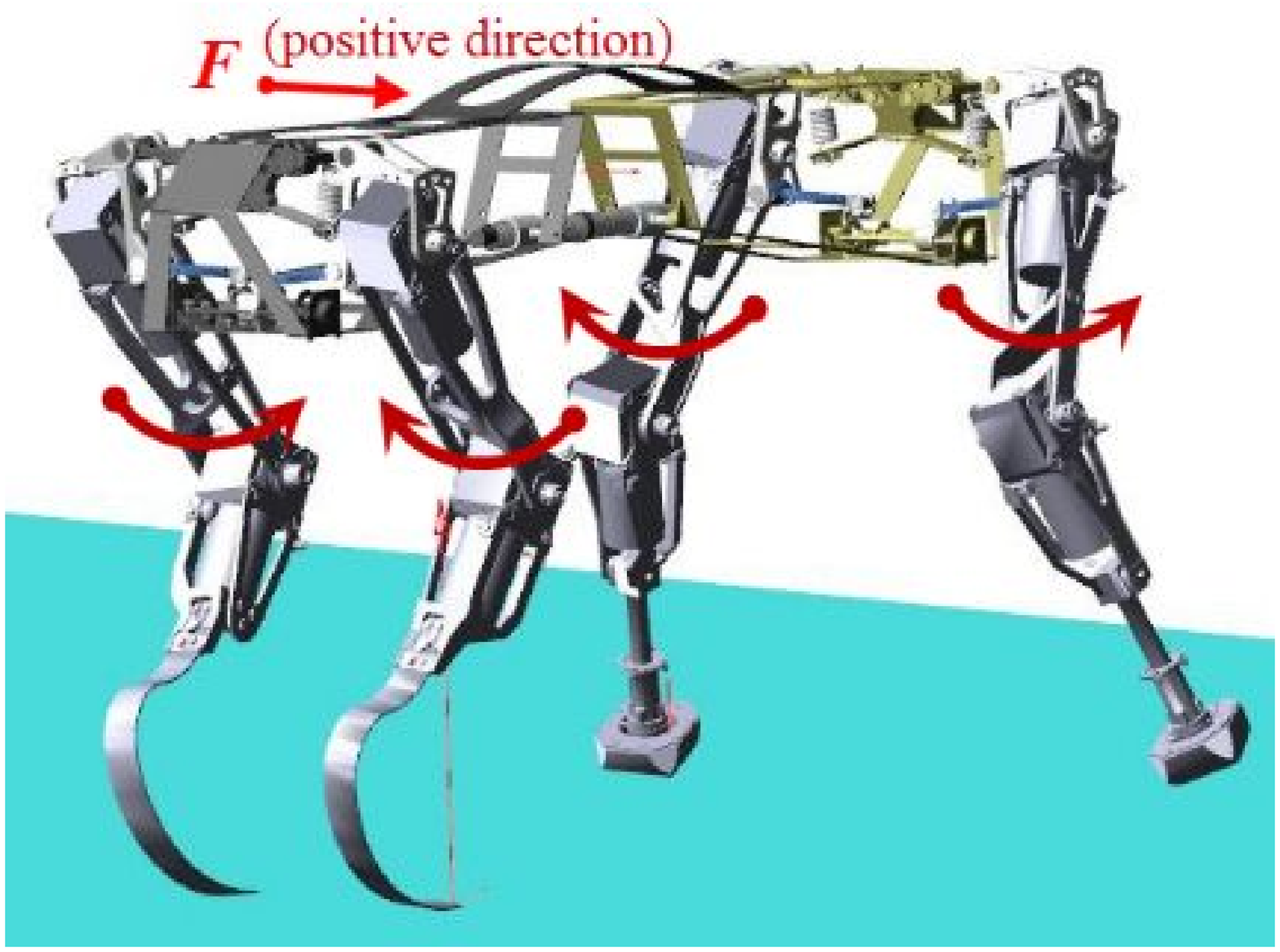}
    \caption{Stepping Direction under Lateral Impact}
    \label{fig:4}
    \end{minipage}
\end{tabular}
\end{figure}

The hip joints of the tetrapod can be regarded as spherical joints which are able to achieve the 3-DOF movement under muscle traction. Tetrapod can timely react to lateral impact by laterally swinging its legs so that the reactive force given by the ground can weaken the lateral acceleration. When the lateral force exceeds the limit, animals will keep laterally stepping until they regain balance. At the same time, animals can effectively prevent the interference between their legs. For Runner, we put forward the lateral stepping control strategy and the following five necessary conditions based on the biologic studying and the simulation to ensure the strategy practicable:
\begin{enumerate}[i]
\item Sliding motion and the lateral force are in the same direction;
\item The phase of the hip joint and the laterally movable joint in the same leg must be consistent, and four hips need to keep a certain phase relationship;
\item No interference occurs between any two legs;
\item During lateral stepping, the feet in the swing phase should not touch the ground too early;
\item During lateral stepping, the robot should be able to choose the proper step length according to different lateral acceleration.
\end{enumerate}
In terms of condition i, it is obvious that only when lateral stepping is in the direction of the lateral impact can the robot reduce the lateral acceleration using the impulse from the ground. In our research, we set up the World Coordinate System (WCS) in Fig.\ref{fig:3} based on the following principles:

\begin{itemize}
\item The left and right directions of Runner are defined from the back view of it;
\item The positive direction of the side direction is from robot's left to its right.
\end{itemize}

Assume that lateral impact is in the positive direction.  To meet the requirement, the leg in the swing phase should step in the positive direction (as is shown in Fig.\ref{fig:4}) and vice versa;

If the lateral movement can be regarded as a kind of walking, it also has swing phase and stance phase, and the swing phase can be gained by the joint effort of the laterally movable joints, hip joints and knee joints. More specifically, the swinging of the hip and knee joints determines the y coordinate of foothold and the height of point, and the swinging of laterally movable joints determines the x coordinate of foothold. In addition, the swing phase of lateral trot must be consistent with that of the longitudinal trot to prevent the legs in stance swing from receiving the laterally moving signal. Meanwhile, Runner's four hip joints have to maintain correct phase relationship to achieve stable walking.

For condition iv, Runner's body will swing around the diagonal line of the supporting legs due to the lateral impact, resulting that the leg in swing phase will touch the ground in advance. This tends to make the robot bear too large load during the moving period or to make it stumble and fall. Therefore, it is essential to ensure the proper leg lifting height so that the leg can be fully off the ground. The following chapters introduce how we adjust the relevant parameters to meet such requirements.

\section{Extended CPG Network}
\subsection{Central Patter Generator}
The CPG control algorithm was derived from biological research. It's inspired by the biological rhythmic movement and uses mathematical methods to describe the a combination controlling the rhythmic movement in the spinal cord, namely the Central Pattern Generator (CPG)\cite{c10}. It uses mathematical model to simulate the interactions between the neurons, which is different from the traditional method to plan the trajectory of the foot. The interaction between the robot and the environment can be achieved by introducing reflexes into the network\cite{c11}. Oscillators, the basic units of CPG, can produce signals with certain phase relationship through their own effect and mutual coupling effect, which acts as the position signals of the robot's joint movement. There are several kinds of oscillators, such as the Matasuoka model, Kimura model and the Hopf model. The former two try to describe the real structure of the neurons\cite{c12}, the Hopf model focuses on the model's output and uses a nonlinear equations to achieve the similar output as the Matasuoka and Kimura models\cite{c13}. We have experimented with all these models and only the Hopf model can quickly tune to oscillate when triggered (this will be detailed described later). The Hopf can be described with the following equations:

\begin{equation}
\label{eq:CPG}
\begin{cases}
\dot{{x_i}}= a(\mu  - r_i^2){x_i} - {\omega _i}{y_i}\\
\dot{{y_i}}= b(\mu  - r_i^2){y_i} + {\omega _i}{x_i} + \sum\limits_{j = 1}^m {{k_{ij}}{y_j}} \\
{r_i} = \sqrt {x_i^2 + y_i^2} \\
{\omega _i} = \frac{{{\omega _{st}}}}{{{e^{ - \tau y}} + 1}} + \frac{{{\omega _{sw}}}}{{{e^{\tau y}} + 1}}\\
{\omega _{st}} = \frac{{1 - \beta }}{\beta }{\omega _{sw}}\\
i = 1,2, \cdots n;j = 1,2, \cdots m
\end{cases}
\end{equation}

\begin{figure}[thpb]
\centering
    \includegraphics[width=2.5in]{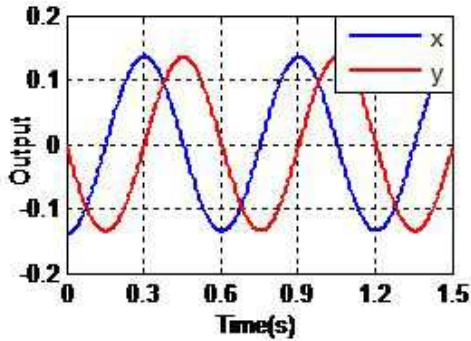}
    \caption{CPG Oscillator Output}
    \label{fig:5}
\end{figure}

The subscript i stands for the serial number of oscillators; n and m represents the number of oscillators and feedback items respectively; x and y are oscillators' output, which are used as Runner Robot's hip and knee phase signal respectively with minor adjustment; \(\mu \) is used to control the amplitude of oscillators. Their relation can be presented by the formula \(A = \sqrt \mu  \); \(\omega \) is closely related to the increasing and decreasing of the output signals. In addition, \({\omega _{st}}\)is stance phase frequency and \({\omega _{sw}}\) is swing phase frequency. They are connected by duty ratio \(\beta \). \(\sum\limits_{j = 1}^4 {{k_{ij}}{y_j}} \) stands for the coupling term between oscillators. \({k_{ij}}\) represents the weight connection of oscillators numbered from i to j and also influences the phase relation between oscillators. When the duty ratio \(\beta \) equals 0.5, the oscillator output curve can be shown as Fig.\ref{fig:5}. The red curve stands for x output, and the blue curve y output. When parameter \(\mu \) is assigned a proper value, the oscillator's x output can be directly used as the phase signal of the hip, meanwhile, the y output, after clipping and amplitude transformation, can function as the knee phase signal of the same leg, and both the x and y output conform to the motion mapping relation of the hip and knee of living organism. As for the lateral movement of the robot, if the x and y direction as well as the structural difference is neglected, there is no essential difference between lateral and longitudinal trot. Thus, the method applied to controlling longitudinal movement also works for lateral movement. Besides, certain adjustment should be introduced to trigger action at the appropriate time and to select the suitable step length and step number, so that the robot can resist the lateral impact of various magnitude.

\subsection{Extended CPG Network}
\begin{figure}[thpb]
\centering
   \includegraphics[width=2in]{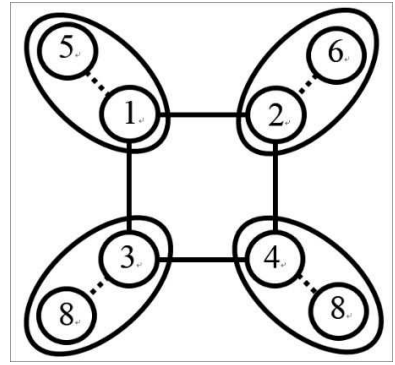}
    \caption{Topological graph of the extended CPG network}
    \label{fig:6}
\end{figure}

We use four oscillators to control the longitudinal trot, and another four to control the LM joints of the robot. Specifically, the four oscillators for LM joints do not work all the time, instead, they will be quickly triggered when the lateral acceleration exceeds a certain value. Their topological relation is shown in Fig.\ref{fig:6}.

The oscillators numbered 1 to 4 control the legs numbered 1 to 4, namely, the left foreleg(LF), right foreleg(RF), right hind leg(RH) and left hind leg(LH). In terms of diagonal gait, the 1 and 3 legs, the 2 and 4 legs are in the same phase respectively, but the two leg groups are in different phase. The 4 LM joints take the same phase with the leg they are attached to. Moreover, the dotted line connecting the hip and LM joint indicates that the LM joints do not work continually.

\subsubsection{Assignment of the connecting weight matrix K}
The connecting weight matrix \({\bf{K}} = {({k_{ij}})_{n \times n}}\) (\(n = 8\))is used to control the phase relation between oscillators. Its assignment conforms to the following principles:
\begin{itemize}
\item each oscillator has its own connecting weight of 0;
\item if two oscillators are in different phase, the value of K is negative (usually -1); By contrast, the value is positive (usually 1) for oscillators in the same phase. 
\item The larger the value is, the greater the connection between oscillators is.
\end{itemize}
In the CPG network with 8 oscillators, we argue that lateral trot is submissive to the longitudinal trot. The matrix K is divided in parts to study its assignment.

\begin{equation}
{\bf{K}} = \left[ {\begin{array}{*{20}{c}}
{{k_{11}}}& \ldots &{{k_{14}}}&{{k_{15}}}& \ldots &{{k_{18}}}\\
 \vdots &{}& \vdots & \vdots &{}& \vdots \\
{{k_{41}}}& \ldots &{{k_{44}}}&{{k_{45}}}& \ldots &{{k_{48}}}\\
{{k_{51}}}& \ldots &{{k_{54}}}&{{k_{55}}}& \ldots &{{k_{58}}}\\
 \vdots &{}& \vdots & \vdots &{}& \vdots \\
{{k_{81}}}& \ldots &{{k_{84}}}&{{k_{85}}}& \ldots &{{k_{88}}}
\end{array}} \right] = \left[ {\begin{array}{*{20}{c}}
{{{\bf{K}}_{\bf{t}}}}&{{{\bf{K}}_{{\bf{ct}}}}}\\
{{{\bf{K}}_{{\bf{tc}}}}}&{{{\bf{K}}_{{\bf{cc}}}}}
\end{array}} \right]
\end{equation}
\({{\bf{K}}_{\bf{t}}}{\bf{,}} {{\bf{K}}_{{\bf{ct}}}}{\bf{,}} {{\bf{K}}_{{\bf{tc}}}}{\bf{,}} {{\bf{K}}_{{\bf{cc}}}}\) are all square matrix of 4x4, and the y output can be described by the following equation:
\begin{equation}
 {{\bf{Y}}^{\bf{T}}} = \left[ {\begin{array}{*{20}{c}}
{{y_1}}&\ldots&{{y_4}}&{{y_5}}&\ldots&{{y_8}}
\end{array}} \right] = {\left[ {\begin{array}{*{20}{c}}
{{{\bf{Y}}_{\bf{1}}}}&{{{\bf{Y}}_{\bf{2}}}}
\end{array}} \right]^{\bf{T}}}
\end{equation}
When the four LM joints' oscillators are triggered, the coupling term of oscillator model is shown as follows:
\begin{equation}
\left[ {\begin{array}{*{20}{c}}
{{{\bf{K}}_{\bf{t}}}{{\bf{Y}}_{\bf{1}}}{\bf{ + }}{{\bf{K}}_{{\bf{ct}}}}{{\bf{Y}}_{\bf{2}}}}\\
{{{\bf{K}}_{{\bf{tc}}}}{{\bf{Y}}_{\bf{1}}}{\bf{ + }}{{\bf{K}}_{{\bf{cc}}}}{{\bf{Y}}_{\bf{2}}}}
\end{array}} \right]
\end{equation}
where,
\begin{itemize}
\item for the 4 hip joints,
\({\left( {{{\bf{K}}_{\bf{t}}}{{\bf{Y}}_{\bf{1}}}} \right)^{\bf{T}}}\) represents the coupling relation of 4 hip joints;
\({\left( {{{\bf{K}}_{{\bf{ct}}}}{{\bf{Y}}_{\bf{2}}}} \right)^{\bf{T}}}\) represents the coupling influence of LM joints on hip joints.
\item for the 4 LM joints,
\({\left( {{{\bf{K}}_{{\bf{tc}}}}{{\bf{Y}}_{\bf{1}}}} \right)^{\bf{T}}}\) shows the coupling influence of hip joints on LM joints
\({\left( {{{\bf{K}}_{{\bf{cc}}}}{{\bf{Y}}_2}} \right)^{\bf{T}}}\) shows the coupling relation between LM joints.
\end{itemize}

Thus, in terms of the gait matrix \({\bf{K = }}\left[ {\begin{array}{*{20}{c}}
{{{\bf{K}}_{\bf{t}}}}&{{{\bf{K}}_{{\bf{ct}}}}}\\
{{{\bf{K}}_{{\bf{tc}}}}}&{{{\bf{K}}_{{\bf{cc}}}}}
\end{array}} \right]\) the function of every sub-block are:
\begin{itemize}
\item \({{\bf{K}}_{\bf{t}}}\) controls the phase relation of 4 hip joints;
\item \({{\bf{K}}_{{\bf{ct}}}}\) controls the coupling influence degree of four LM joints on hip joints;
\item \({{\bf{K}}_{{\bf{tc}}}}\) controls the coupling influence degree of four hip joints on LM joints;
\item \({{\bf{K}}_{{\bf{cc}}}}\) controls the phase relation of four LM joints.
\end{itemize}
Considering the above-mentioned two amplitude principles, the matrix \({\bf{K}}\) can be assigned in the following way:
\begin{itemize}
\item To meet the condition ii, legs of 1, 3 and 2, 4 are respectively in the same phase and the two leg groups are in different phase:
\begin{equation}
{{\bf{K}}_{\bf{t}}} = \left[ {\begin{array}{*{20}{c}}
0&{ - 1}&1&{ - 1}\\
{ - 1}&0&{ - 1}&1\\
1&{ - 1}&0&{ - 1}\\
{ - 1}&1&{ - 1}&0
\end{array}} \right]
\end{equation}
\item To ensure that the LM joints keep the same phase with their corresponding hip joints, the coupling coefficient should be positive. The simulation results in Fig.\ref{fig:7} showed that when the coupling coefficient is 1, the signal of oscillators in LM joints fails to keep the same phase with the hip joints at the beginning, but succeeds after several cycles. When the coupling coefficient equals 5, the signal can keep pace with that of hip joints in 1/4 cycle after the LM joints starts oscillating. Thus, the mathematical relation about matrix \({{\bf{K}}_{{\bf{tc}}}}\) are presented in the following equation. \(\lambda \) ranges from 3 to 5. It should not be too large, otherwise the initial stability and the amplitude of the signal will be perturbed.
\begin{figure}[thpb]
      \centering
    \begin{tabular}{cc}
    \begin{minipage}[t]{1.5in}
    \includegraphics[width=1.5in]{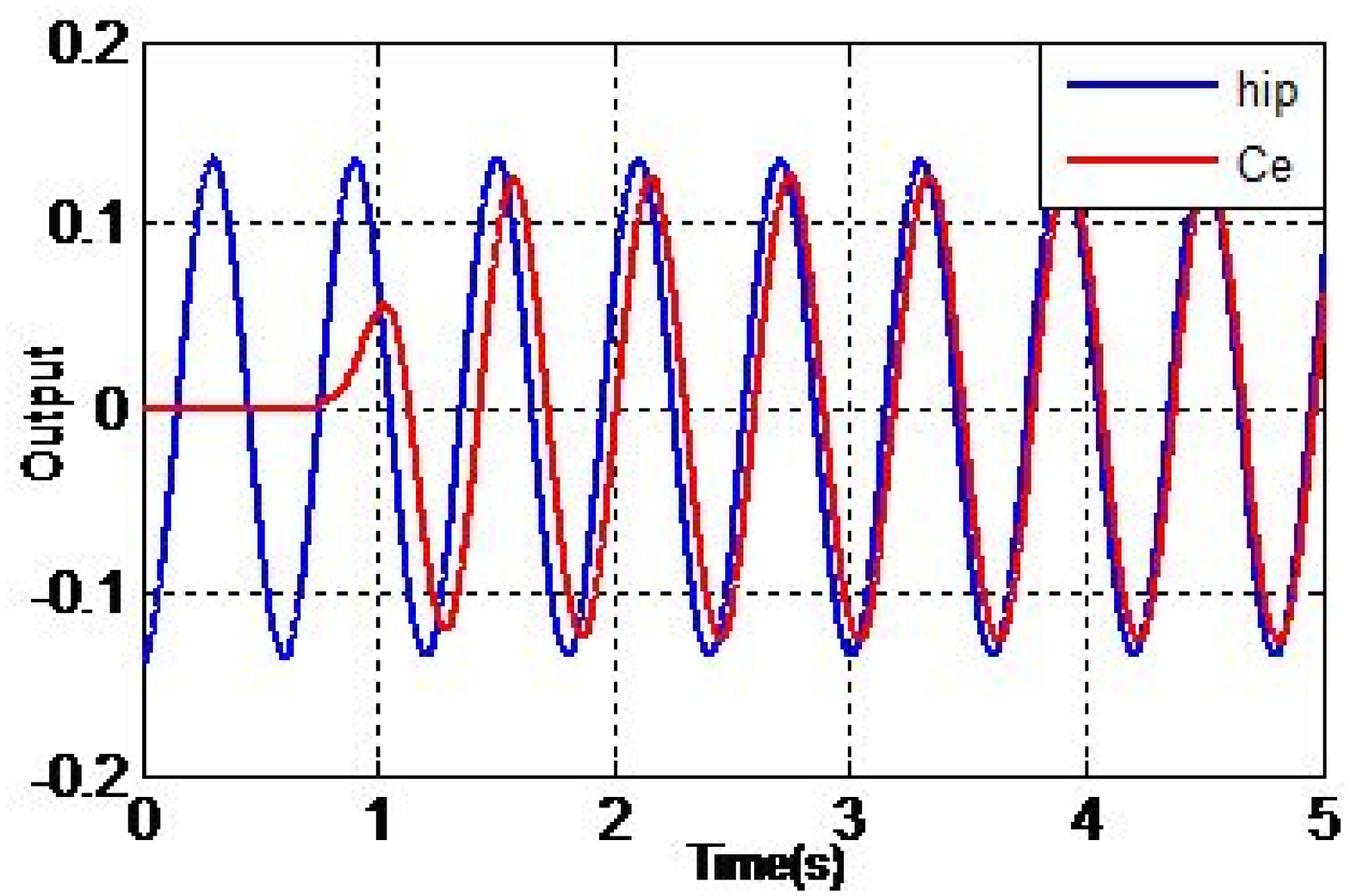}
    \end{minipage}
    \begin{minipage}[t]{1.5in}
   \includegraphics[width=1.5in]{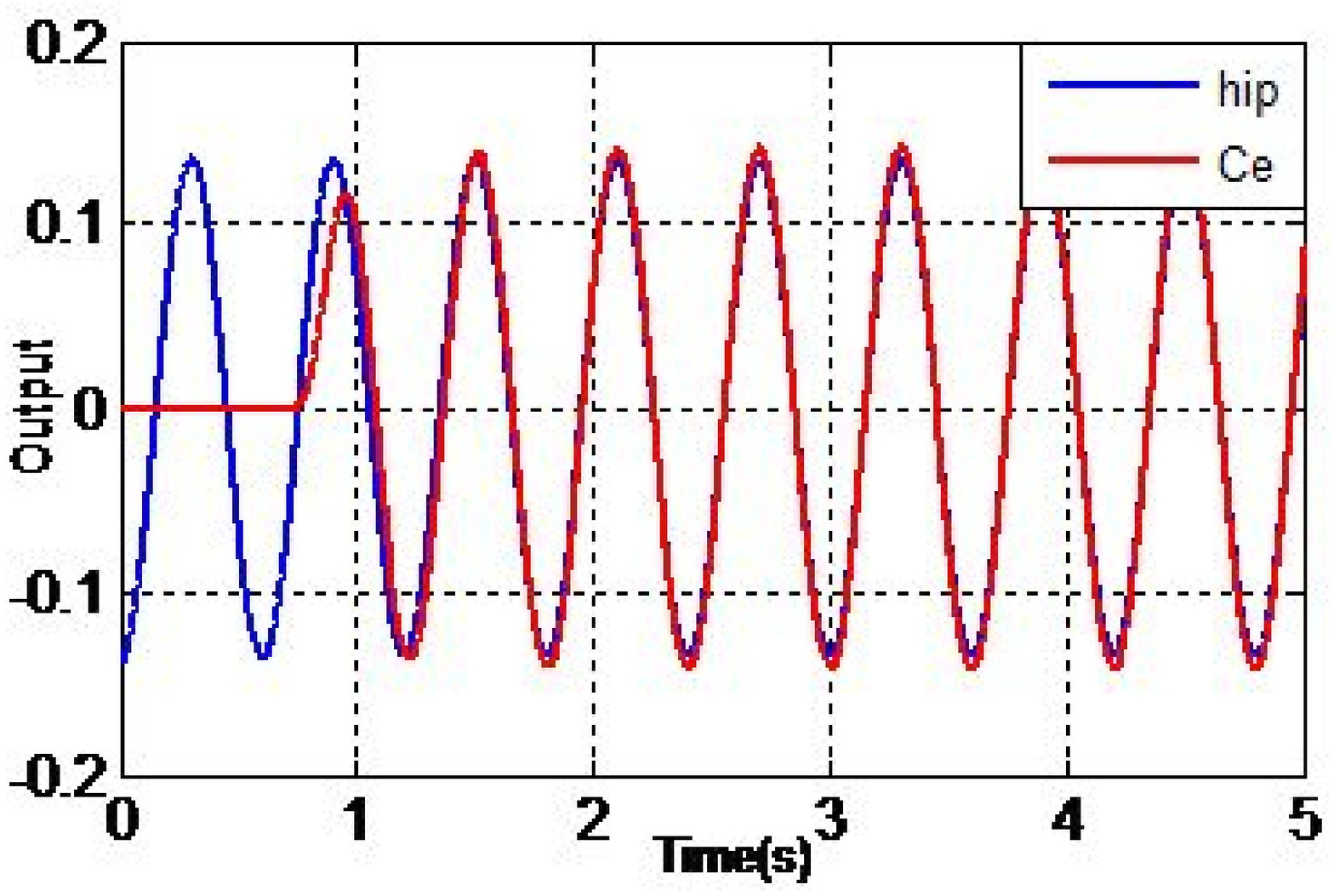}
    \end{minipage}
    \end{tabular}
    \caption{The output when coupling coefficient is 1(left) and 5 (right)}
    \label{fig:7}
\end{figure}
\begin{equation}
   {{\bf{K}}_{{\bf{tc}}}} = \lambda {\bf{I}},\lambda  = 1,2,3,4 \ldots
\end{equation}
\item Since the lateral trot is submissive to the longitudinal trot, the influence between LM joints and hips should be one-directional, thus \({{\bf{K}}_{{\bf{ct}}}} = {\bf{0}}\).
\item The matrix \({{\bf{K}}_{{\bf{cc}}}}\) controls the coupling effect of the 4 LM joints. Because the LM joints have already retain a particular phase relation with their corresponding hip joints through the matrix \({{\bf{K}}_{{\bf{tc}}}}\), we can assume \({{\bf{K}}_{{\bf{tc}}}}{\bf{ = 0}}\).
\end{itemize}
Above all, the gait matrix \({\bf{K}} = \left[ {\begin{array}{*{20}{c}}
{{{\bf{K}}_{\bf{t}}}}&{\bf{0}}\\
{\lambda {\bf{I}}}&{\bf{0}}
\end{array}} \right]\).

\begin{figure}[thpb]
      \centering
  \includegraphics[width=2.5in]{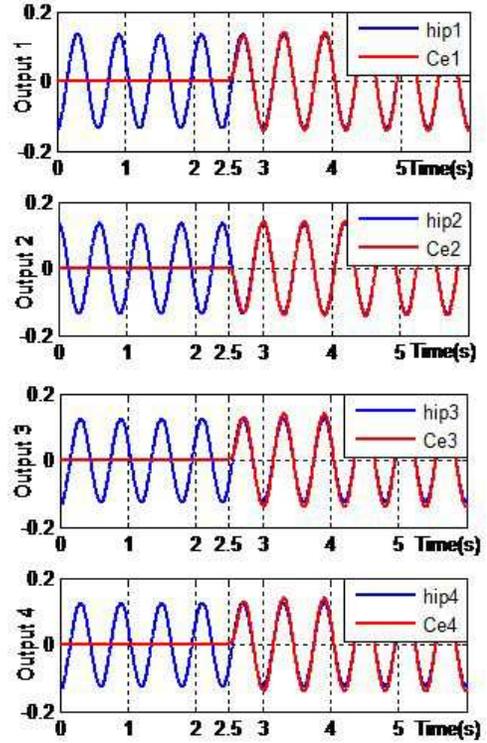}
\caption{Numerical simulation results from Simulink}
\label{fig:9}       
\end{figure}

The rest parameters can be determined according to \cite{c5}. First we neglect the amplitude of the LM joints. Suppose that the lateral impact is received at \(t = 2.5s\), and \(\lambda \) equals 4, the network output is shown in Fig.\ref{fig:9}. As we can see, the output of LM joints, the red curve, can start oscillating immediately at the triggering time \(t = 2.5s\). It can remain the same phase as the curve of hip joints, totally consistent with the requirement of condition i and ii.

\subsubsection{Amplitude of hip joints}
If the hip joint does not swing or only swing with small amplitude, the legs are liable to interference when one leg swings inward. Therefore the signal has to reach the minimum amplitude. This is related to the leg structure, foot volume and longitudinal moving speed. For our robot, the minimum amplitude of hip joint equals 0.174rad.

While Runner is trotting longitudinally, the hip swing amplitude can be identified according to Fig.\ref{fig:11}:
\begin{figure}[thpb]
      \centering
      \includegraphics[width=1.5in]{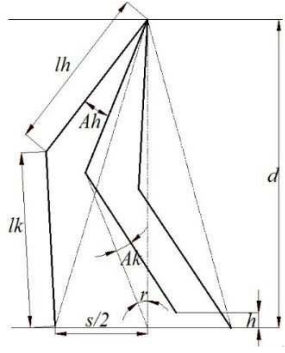}
    \caption{Swing phase of the Runner}
    \label{fig:11}
\end{figure}
\begin{equation}
\left\{ \begin{array}{l}
{A_h} = \arctan \frac{{{\raise0.5ex\hbox{$\scriptstyle s$}
\kern-0.1em/\kern-0.15em
\lower0.25ex\hbox{$\scriptstyle 2$}}}}{d}\\
s = \frac{{vT}}{2}
\end{array} \right.
\end{equation}
where the v stands for the longitudinal speed and T is the period of one gait.
The \(\mu\) in Eq. \ref{eq:CPG} is adapted as:
\begin{equation}
\mu  = \left\{ \begin{array}{l}
{\mu _0} \times {\left( {\frac{{{A_{h\min }}}}{{{A_{h0}}}}} \right)^2},if {A_{h0}} < {A_{h\min }}\\
{\mu _0},if {A_{h0}} \ge {A_{h\min }}
\end{array} \right.
\end{equation}
where \({A_h}_0\) represents the hip joint amplitude of longitudinal trot.

\subsubsection{Amplitude of knee joints}
When Runner receives the lateral impact, it swings around the diagonal line of the supporting legs. As a result, the leg in swing phase touches the ground earlier. Given that the swing of the knee joint is the direct factor affecting the leg lifting height, the swing amplitude of knee joint should be increased properly to make sure that the leg won't touch the ground too early. The angular acceleration can be measured by the angular acceleration sensor carried by the robot, and the time of swing phase period is \({t_{sw}} = 0.3s\), so the descent height of the foot point is:
\begin{equation}
h_c = \frac{1}{2}\beta {t_{sw}}^2{l_c}
\end{equation}
\begin{figure}[thpb]
\centering
  \includegraphics[width=3in]{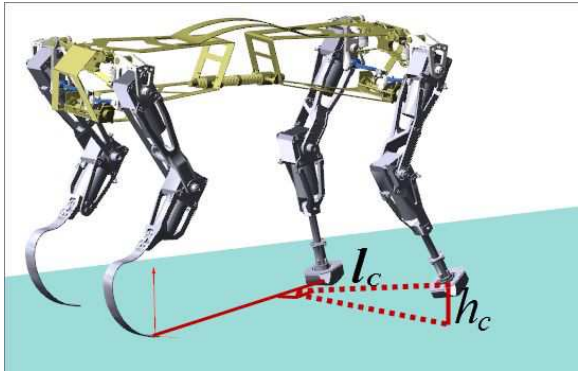}
\caption{Runner's swinging by the supporting axis}
\label{fig:12}       
\end{figure}

\({l_c}\) is the distance between the foot endpoint of the leg in swing phase and the supporting line, and \({h_c}\) is the minimum lifting height, as can be seen in Fig.\ref{fig:12}. The \({h_c}\) is calculated every time the foot is going to lift up. According to Fig.\ref{fig:11}, the minimum amplitude of knee joint is:
\begin{equation}
{A_k}_{\min } = \arccos \frac{{d\cos {A_h} - {l_h}\cos \theta  - {h_c}}}{{{l_k}}} - \gamma 
\end{equation}
In the lateral movement,\({A_k} = \max \left( {{A_k}_{\min },{A_{k0}}} \right)\).

The y output of oscillators is adjusted as follows to get the knee joint signal
\begin{equation}
kne{e_i} = \left\{ \begin{array}{l}
sign\frac{{Ak}}{{Ah}}{y_i},{y_i} \le 0\\
0,{y_i} \le 0
\end{array} \right.
, sign = \left\{ \begin{array}{l}
 + 1, {elbow}\\
 - 1, {knee}
\end{array} \right.
\end{equation}

\section{ZMP and inverted pendulum model}
To meet the condition v, Runner should be able to determine the lateral step length under various magnitude of lateral impact. Modeling the robot as a linear inverted pendulum model\cite{c15}, we developed the Zero Moment Point (ZMP)\cite{c14} based algorithm for the calculation of lateral step length, which is composed of two parts: the positioning of the ZMP and the planning of the position of Center of Gravity (CoG).

\subsection{Calculation of ZMP}
According to \cite{c14}, \cite{c16}, the ZMP is defined as a point on the supporting plane. The horizontal component of the resultant moment of the inertial force and the gravity is zero w.r.t. this point. When the robot is motionless, the ZMP is the vertical projecting point of the CoG on the supporting plane; when the robot is dynamic, ZMP is the projecting point of the CoG in the direction of the resultant force of the inertial force and the gravity, on the supporting plane. Since we focus on the lateral movement, so only the ZMP in robot's lateral direction is needed. Runner Robot's ZMP in x direction can be determined by:
\begin{equation}
{x_{ZMP}} = \frac{{\sum\limits_i {{m_i}\left( {{{\ddot z}_i} + g} \right){x_i} - \sum\limits_i {{m_i}\left( {{{\ddot x}_i} + g} \right){z_i}} } }}{{\sum\limits_i {{m_i}\left( {{{\ddot z}_i} + g} \right)} }}
\end{equation}
where \({m_i}\) is the mass of each part of every robot, \({x_i}\) and \({z_i}\) represent the centroid coordinate of each part, \(g\) stands for the gravitational acceleration. The direction of \(g\) is the negative direction of z.
\begin{figure}[thpb]
\centering
  \includegraphics[width=3in]{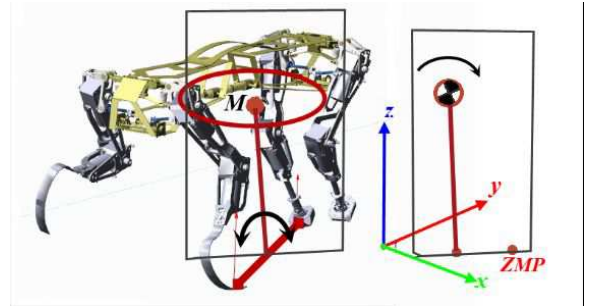}
\caption{the inverted pendulum model}
\label{fig:13}       
\end{figure}

\subsection{Planning of CoG position}
The linear inverted pendulum model is introduced here. Suppose that the mass of the robot centers in the CoG, with the ZMP as the supporting point of the inverted pendulum, we can simplify the robot to a two-dimension inverted pendulum model, and the swinging plane is perpendicular to line fixed by the supporting legs, as is shown in Fig.\ref{fig:13}. Assume that the coordinate of CoG is \(({x_g},{y_g},{z_g})\) , then relation of ZMP and CoG can be described as follows \cite{c17}:

\begin{equation}
\left[ {\begin{array}{*{20}{c}}
{{x_{ZMP}}}\\
{{y_{ZMP}}}
\end{array}} \right] = \left[ {\begin{array}{*{20}{c}}
{{x_g}}\\
{{y_g}}
\end{array}} \right] - \frac{{{z_g}}}{g}\left[ {\begin{array}{*{20}{c}}
{{{\ddot x}_g}}\\
{{{\ddot y}_g}}
\end{array}} \right]\
\end{equation}
Take the x coordinate for example, the calculating formula is
\begin{equation}
\left[ {\begin{array}{*{20}{c}}
{{x_g}\left( t \right)}\\
{{{\dot x}_g}\left( t \right)}
\end{array}} \right] = {\bf{T}}\left( t \right)\left[ {\begin{array}{*{20}{c}}
{{x_{{g_0}}}}\\
{{{\dot x}_{{g_0}}}}
\end{array}} \right] + \left( {{\bf{I}} - {\bf{T}}\left( t \right)} \right)\left[ {\begin{array}{*{20}{c}}
{{x_{ZMP}}}\\
0
\end{array}} \right]
\label{eq:ZMP}
\end{equation}
with,
\[{\bf{T}}\left( t \right) = \left[ {\begin{array}{*{20}{c}}
{\cosh \left( {qt} \right)}&{\frac{1}{q}\sinh \left( {qt} \right)}\\
{q\sinh \left( {qt} \right)}&{\cosh \left( {qt} \right)}
\end{array}} \right]\]
\[q = \sqrt {\frac{g}{{{z_g}}}} \]
where \({x_{{g_0}}}\) and \({\dot x_{{g_0}}}\) represent the initial coordinate and speed of the CoG in x direction respectively. \({\bf{I}}\) is a unit matrix of \(2 \times 2\). Let the state equation of the inverted pendulum system be \({\bf{D}}\left( t \right) = {\left[ {\begin{array}{*{20}{c}}
{x\left( t \right)}&{\dot x\left( t \right)}
\end{array}} \right]^{\bf{T}}}\). When the position of ZMP is \({{\bf{C}}^{\bf{T}}} = \left[ {\begin{array}{*{20}{c}}
{{x_{ZMP}}}&0
\end{array}} \right]\), the Eq.\ref{eq:ZMP} can formulated as:
\begin{equation}
{\bf{D}}\left( t \right) = {\bf{T}}\left( t \right){\bf{D}}\left( 0 \right) + \left( {{\bf{I}} - {\bf{T}}\left( t \right)} \right){\bf{C}}
\end{equation}
As initial the position of ZMP is known, the CoG position which will maintain robot's stability the next moment can also be calculated step by step and the planning of CoG position can be completed. With coordinate transformation, we can obtain the foothold in x direction\cite{c18}; considering the relation between the stepping point in x direction and LM joint position indicated in Eq.\ref{eq:footLat}, we can get the LM joints' amplitude.

It should be noted that the calculation of the knee and hip joint amplitude is executed when the joint rotative angle is 0, likewise, the alteration of step length and rotative angle amplitude of knee joint should be executed under the same condition. In addition, Runner's lateral step length is limited by its mechanical structure. When it receives lateral impact of large magnitude, Runner needs to make several lateral steps to retain stability.

\begin{figure}[thpb]
\centering
  \includegraphics[width=3.5in]{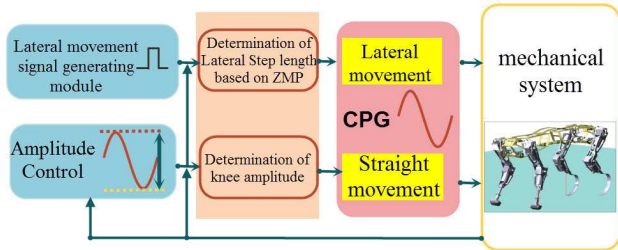}
\caption{method for lateral dynamic stability maintaining}
\label{fig:14}       
\end{figure}

Till now, we have established a complete method for lateral stability maintaining (in Fig.\ref{fig:14}). This method can meet the requirements of the five conditions mentioned in the former part very well. The combined simulation based on Matlab/Simulink and Adams is applied to verify the feasibility and effectiveness of the method.

\section{Simulation Results}
\subsection{Virtual Prototype in Adams}
\begin{figure}[thpb]
\centering
    \begin{tabular}{cc}
    \begin{minipage}[t]{1.5in}
    \includegraphics[width=1.5in]{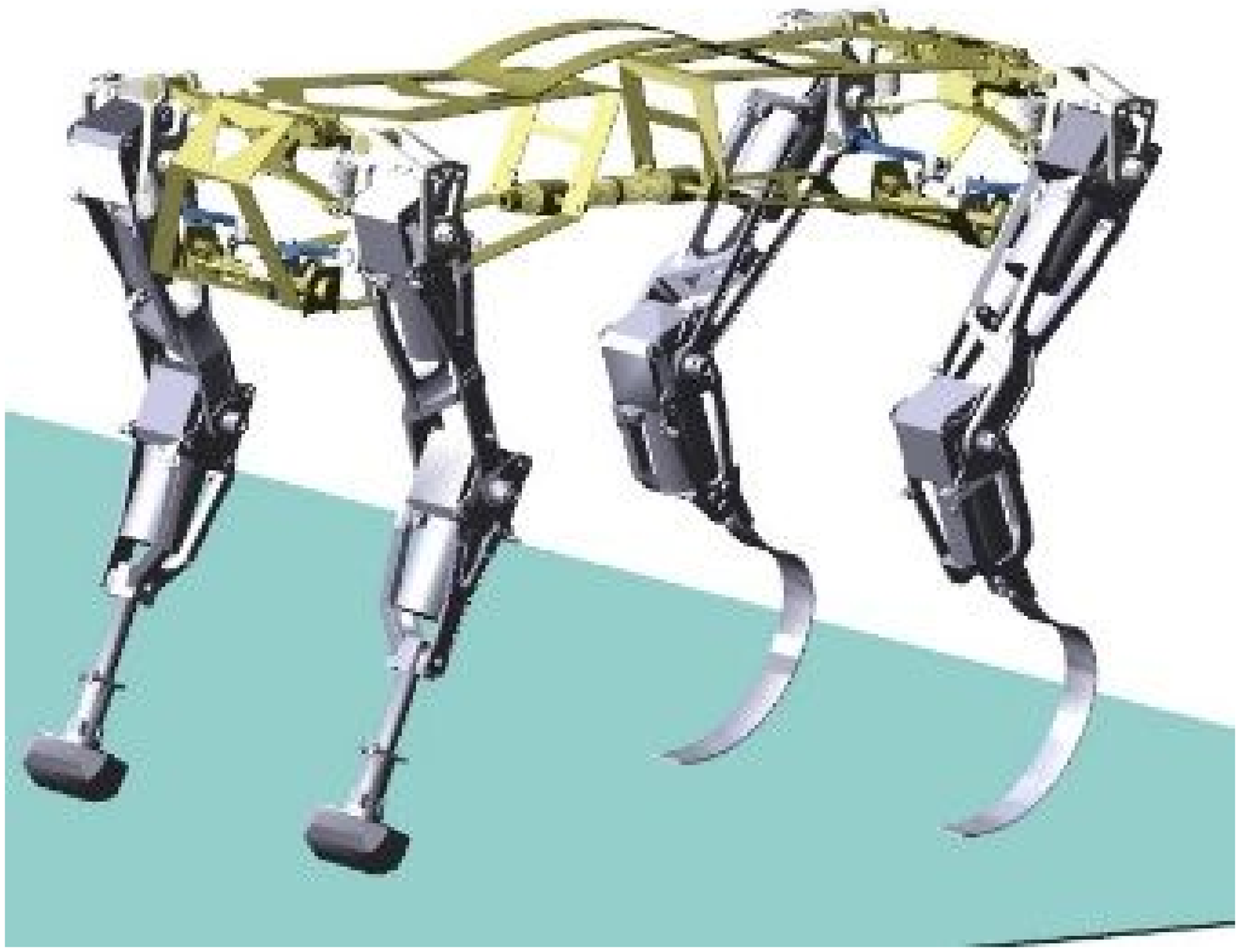}
    \caption{The imported model}
    \label{fig:15}
    \end{minipage}
    \begin{minipage}[t]{1.5in}
   \includegraphics[width=1.5in]{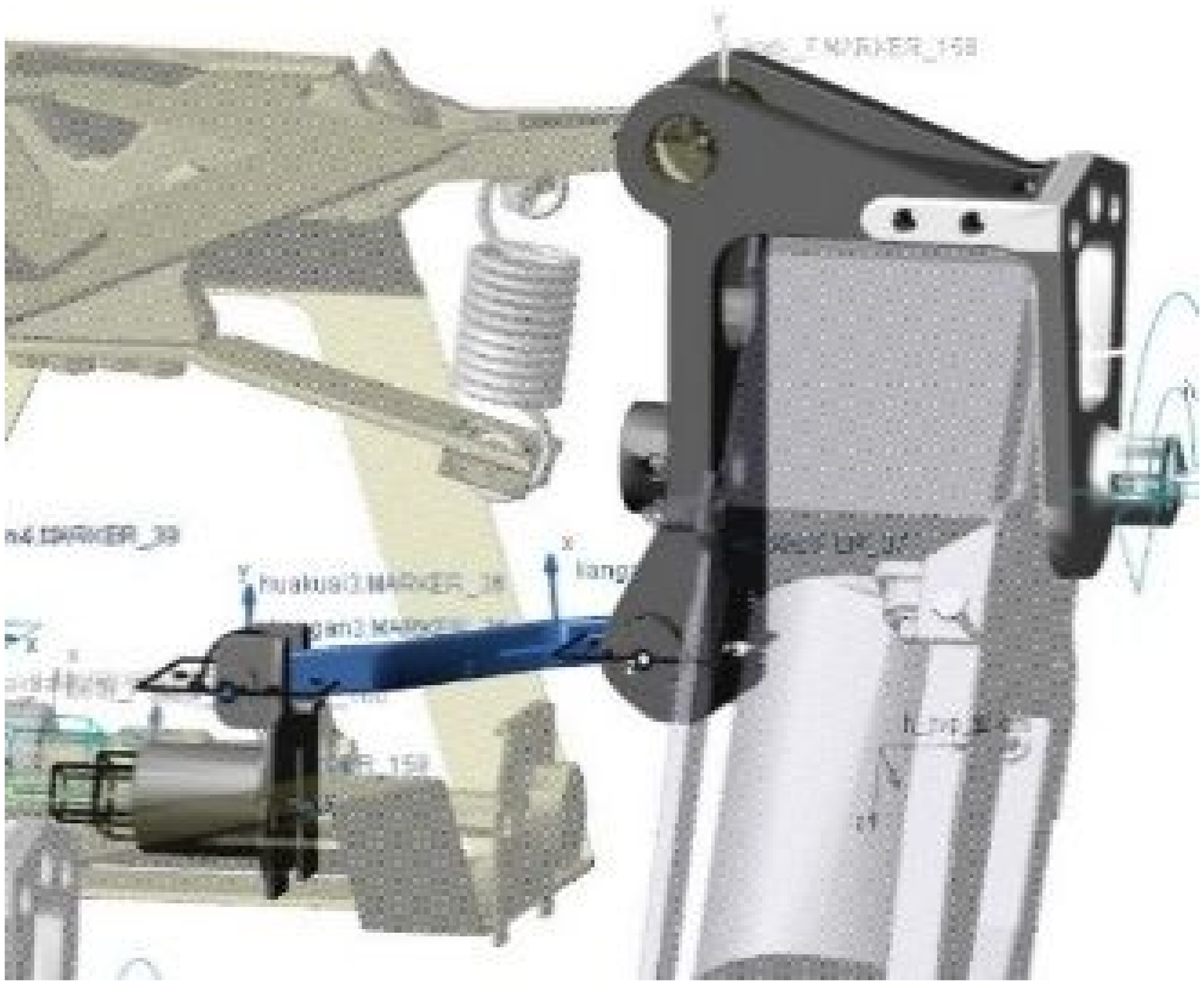}
    \caption{kinematic pair and drive parts}
    \label{fig:16}
    \end{minipage}
\end{tabular}
\end{figure}

We use the Solidworks to assemble the individual parts of Runner Robot and import the assembly model into the Adams in Parasolid format, as is shown in Figure\ref{fig:15}. The contact between 4 feet and the ground is defined. The kinematic pairs between parts and drives are set up as Fig.\ref{fig:16} shows.

\subsection{Control model in Matlab/simulink}
The control model is built in Matlab/Simulink and communicates with Adams model with Adams/Control module. The control system adopts the modular design, which consists of longitudinal trot control module, lateral movement signal generating module, amplitude calculating module.

\subsection{Experiments}
\begin{figure}[thpb]
\centering
  \includegraphics[width=3.5in]{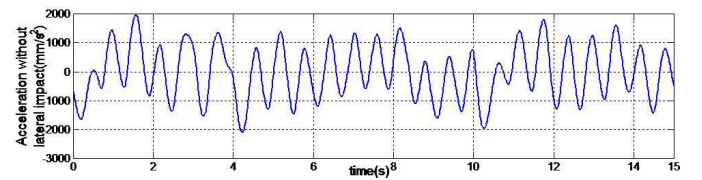}
\caption{change of acceleration without lateral impact}
\label{fig:17}       
\end{figure}

First we conduct the simulation without lateral impact so as to get the acceleration of normal walking, determining the threshold value of lateral stepping reflex for following simulation. The acceleration of the body is measured in the simulation and then imported into Adams/Processor module for post processing. The acceleration is filtered with low-pass Butterworth filter with the frequency of 1.67Hz (the same as walking frequency), as is shown in Figure\ref{fig:17}.

\begin{figure}[thpb]
\centering
\subfigure[normal walking]{
\label{fig:18a}
\includegraphics[width=1.5in]{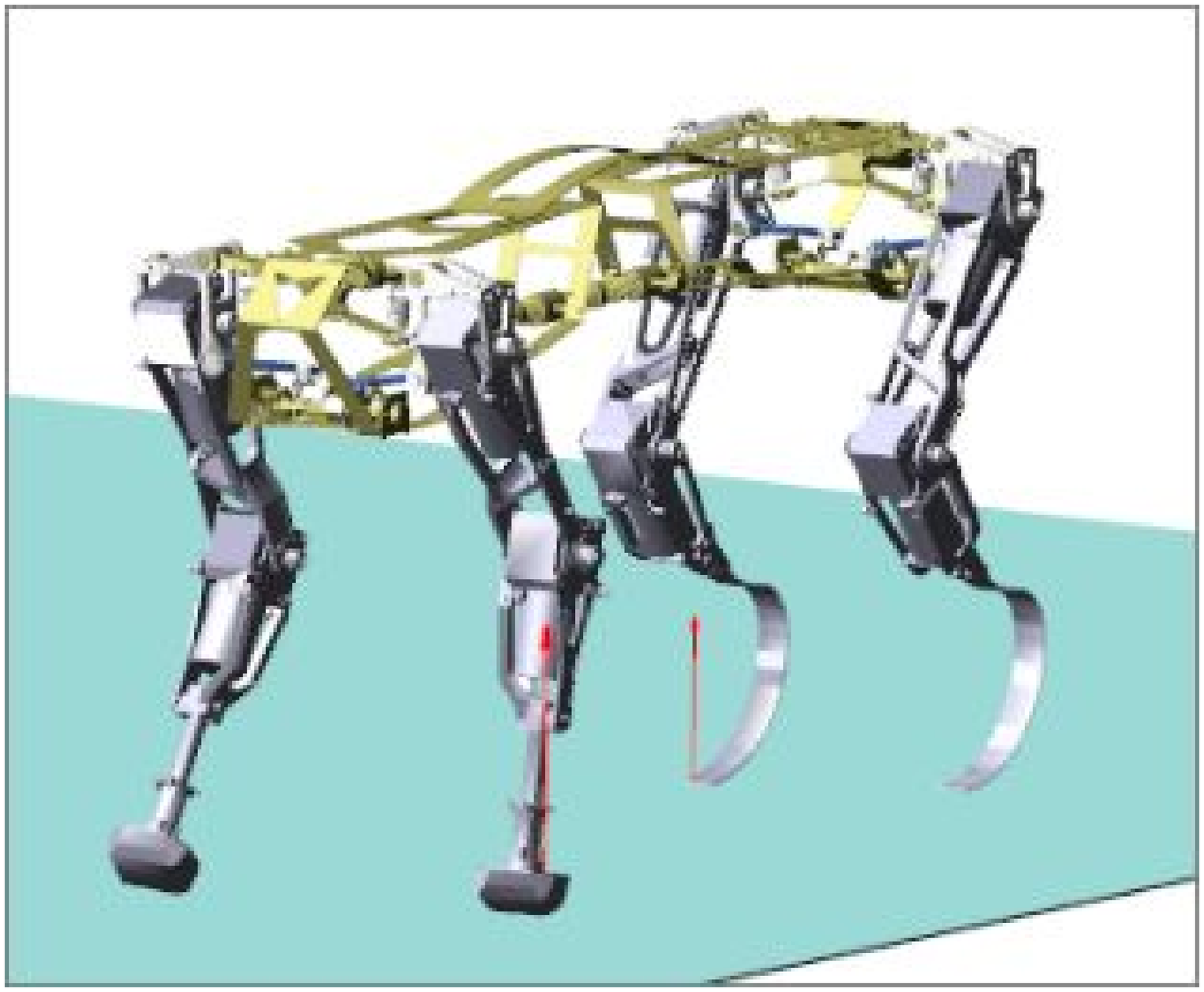}}
\subfigure[lateral impact]{
\label{fig:18b}
\includegraphics[width=1.5in]{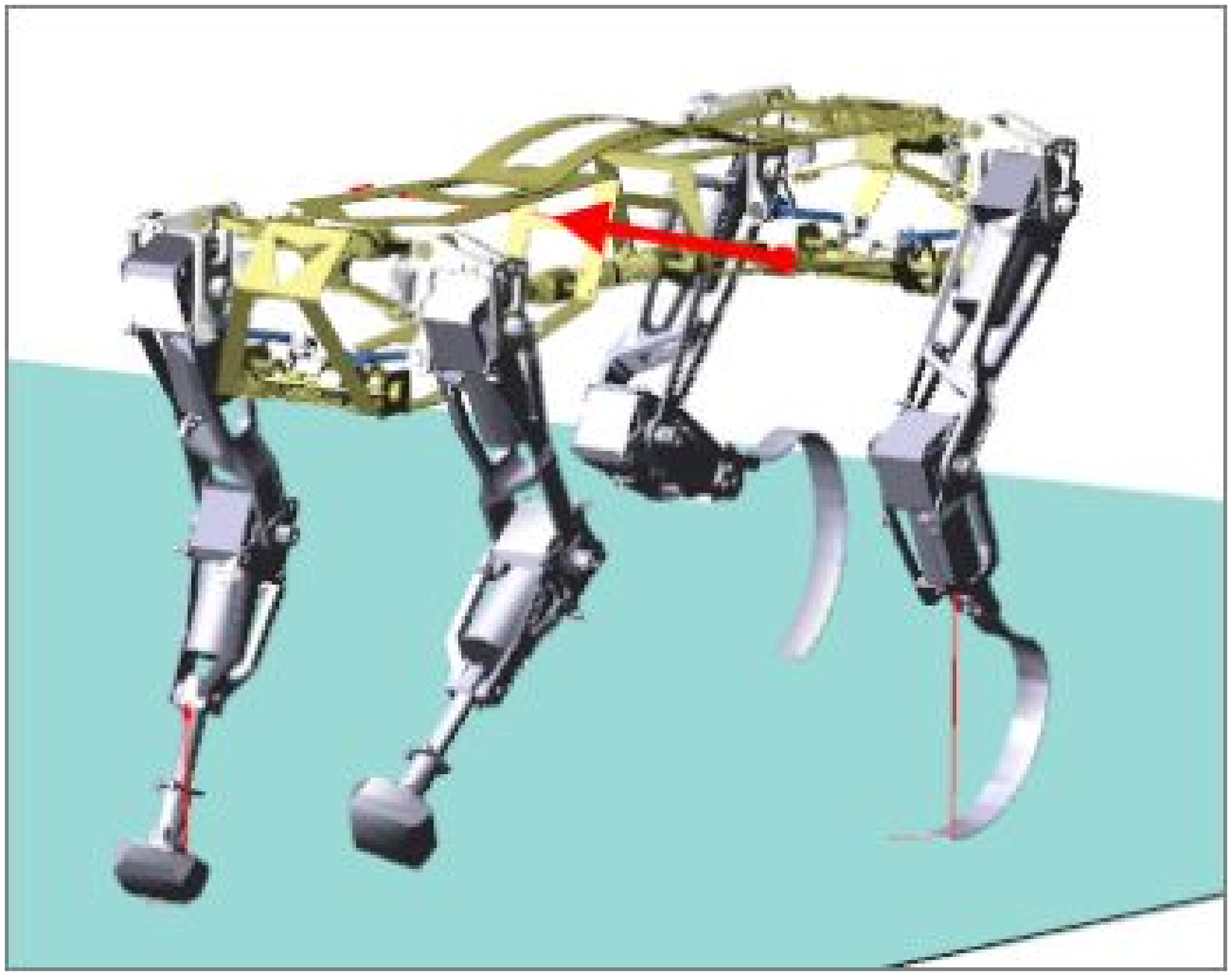}}
\subfigure[lateral stepping]{
\label{fig:18c}
\includegraphics[width=1.5in]{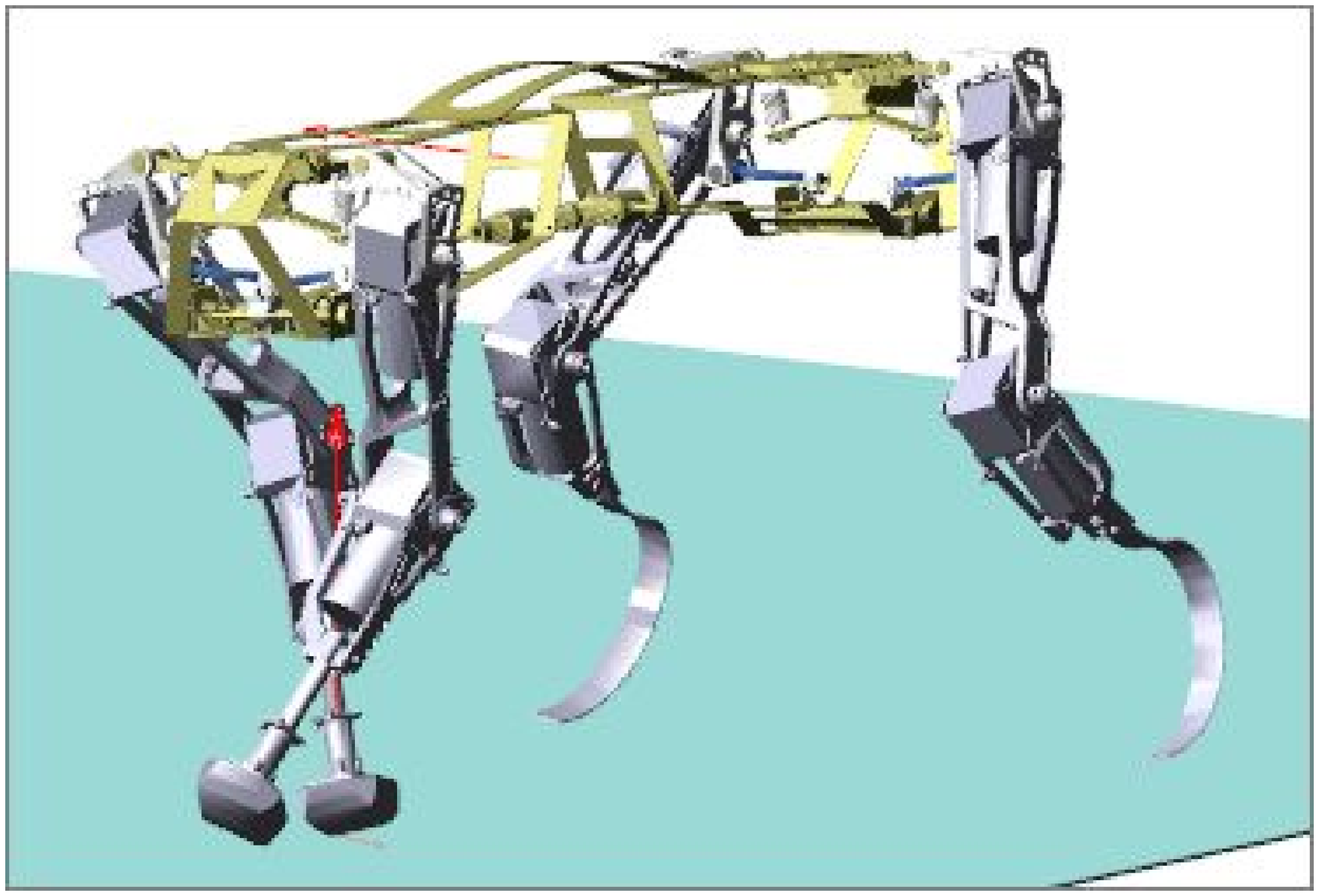}}
\subfigure[return to normal]{
\label{fig:18d}
\includegraphics[width=1.5in]{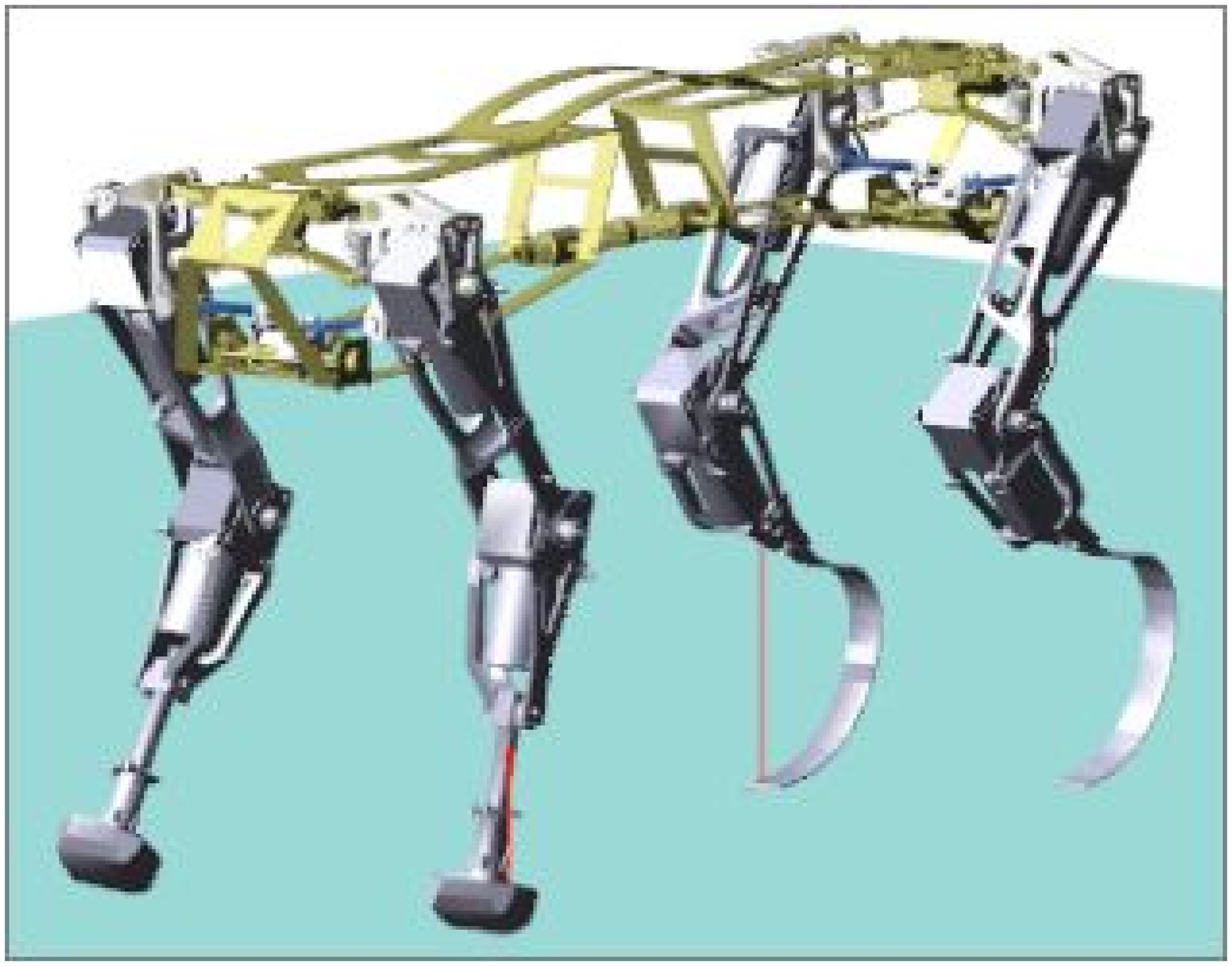}}
\caption{the recording of motion simulation of Runner Robot}
\label{fig:18}
\end{figure}

It can be seen from Figure\ref{fig:17} that the acceleration is kept within \(2000mm/{s^2}\), so the threshold is set as \(2500mm/{s^2}\). The starting time and magnitude of the lateral impact are set as \(t = 2.5s\) and 220N (the value is determined by the weight of Runner Robot, about 22kg) with duration of 0.2s. The result is shown in Figure\ref{fig:18}.
(a) is the video screenshot of robot's normal walking in diagonal gait; the red arrow in (b) indicates the lateral impact on the CoG point; (c) shows the robot's lateral stepping under the lateral impact and robot's four legs are keeping the same phase with the hip joint; in screenshot (d), Runner stops laterally stepping when the lateral acceleration regains to normal.

\subsection{Analysis}
\begin{figure}[thpb]
\centering
  \includegraphics[width=3.5in]{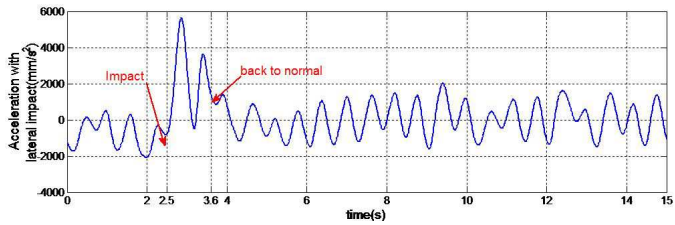}
\caption{the alteration of acceleration before and after the lateral impact}
\label{fig:19}
\end{figure}

The lateral acceleration is clearly presented in Fig.\ref{fig:19}. At \(t = 2.5s\), the robot receives the lateral impact and its acceleration changed abruptly and reached \(5633.1mm/{s^2}\), triggering the lateral stepping. With the help of the lateral stepping, the lateral acceleration return to normal in two cycles.

\begin{figure}[thpb]
\centering
  \includegraphics[width=3in]{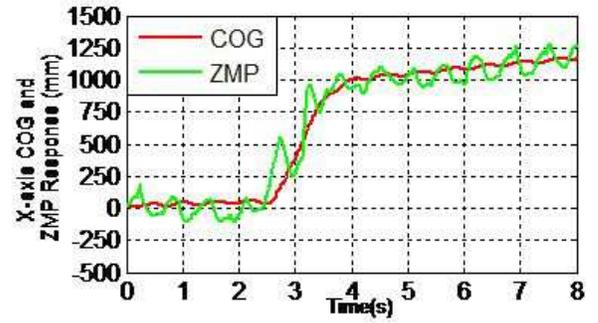}
\caption{x coordinates of CoG and ZMP}
\label{fig:20}       
\end{figure}

The x coordinates of CoG and ZMP are shown in Fig.\ref{fig:20}, in which the red solid line represents x coordinate of CoG, and the green solid line ZMP. Due to the existence of lateral acceleration during normal trot, ZMP's x coordinates fluctuates around 0. The coordinates' fluctuation CoG is tiny at normal condition. At \(t = 2.5s\), the lateral impact 220N is exerted on the robot, resulting in the rapid enlargement of ZMP's x coordinates. Through lateral stepping, the robot tries to make sure that the actual CoG coincides with the planned CoG. Altogether the robot completed two lateral stepping, in which the counterforce provided by the ground has generated effect on robot's lateral acceleration and further brought about the big fluctuation of ZMP's coordinates. At \(t = 3.6s\), the acceleration is within the threshold value again, followed by the termination of lateral stepping.

\begin{figure}[thpb]
\centering
  \includegraphics[width=3in]{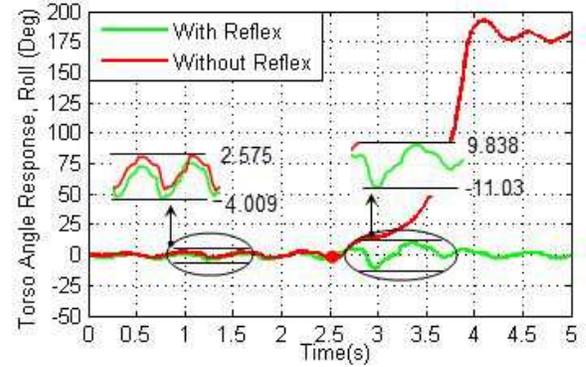}
\caption{roll angle}
\label{fig:21}       
\end{figure}

\begin{figure}[thpb]
\centering
\subfigure{
\includegraphics[width=1.5in]{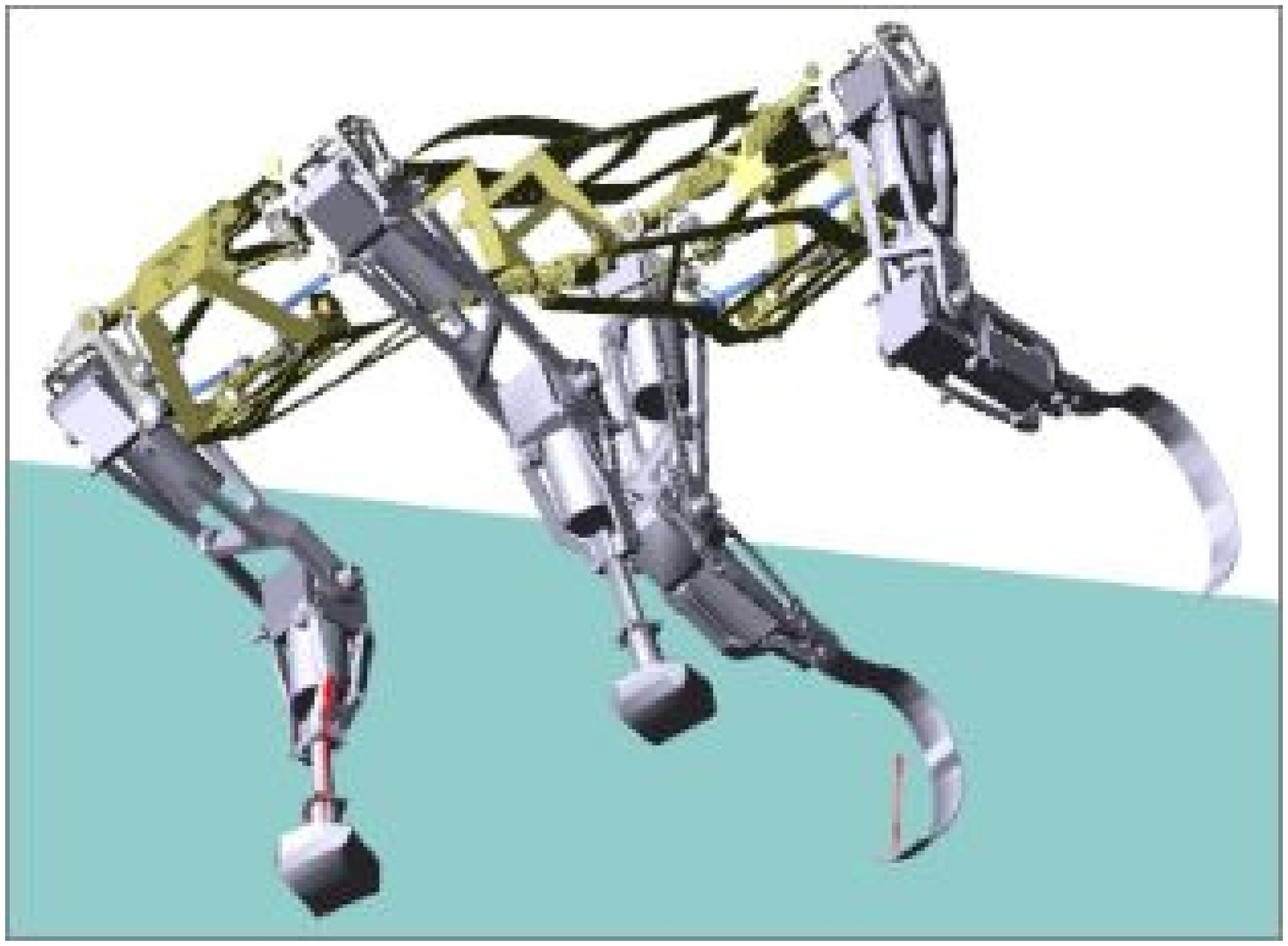}}
\subfigure{
\includegraphics[width=1.5in]{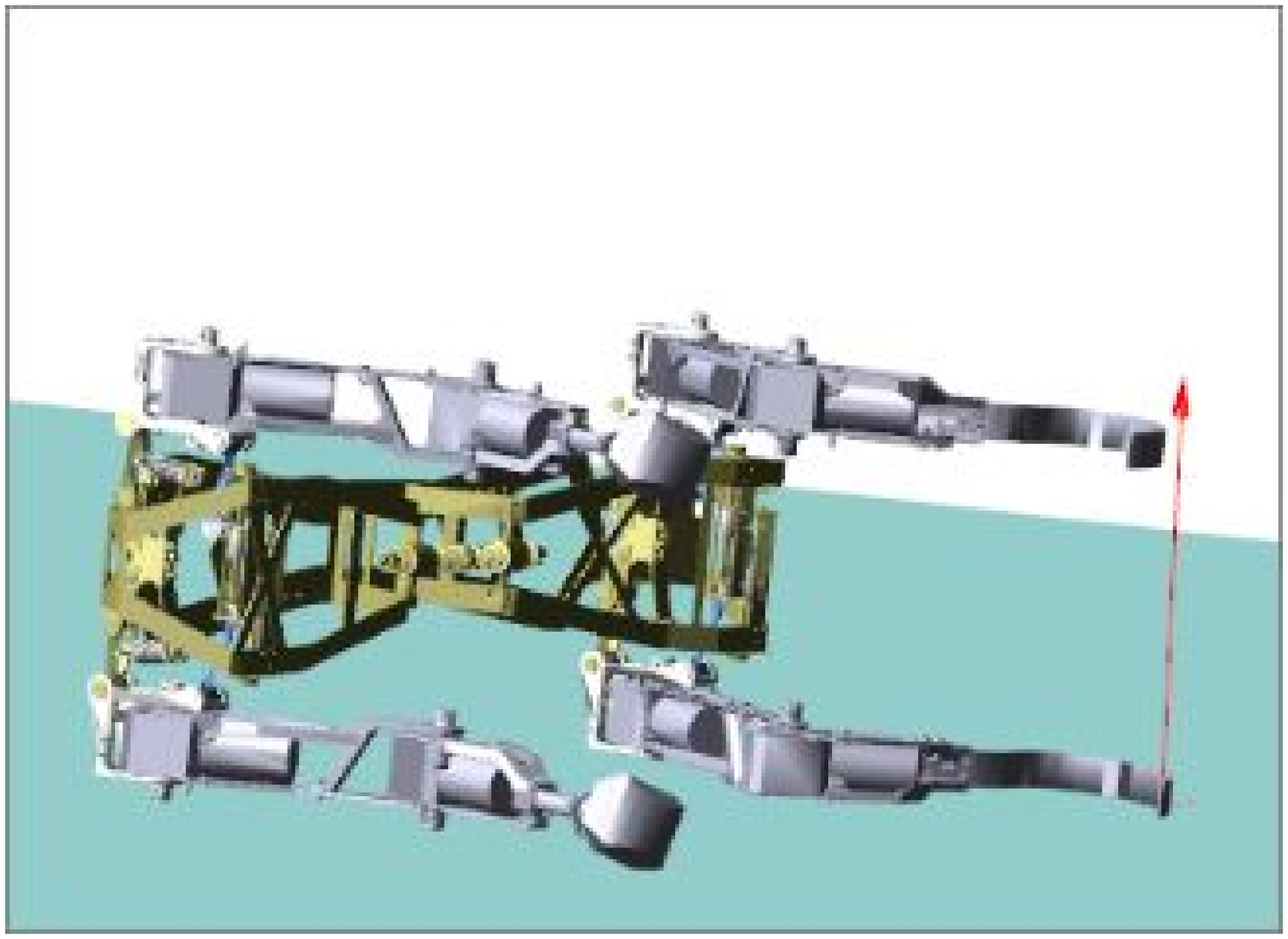}}
\subfigure{
\includegraphics[width=1.5in]{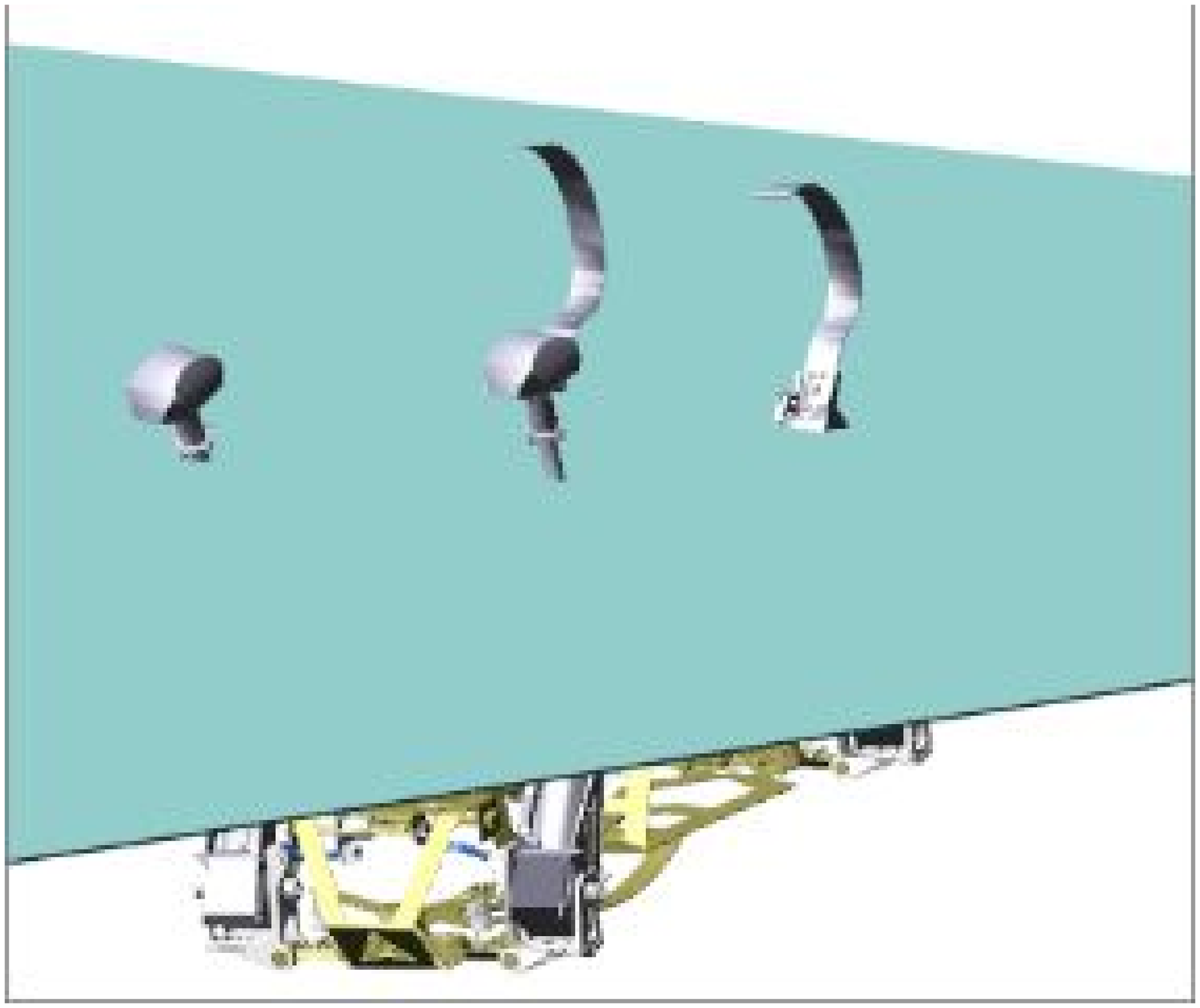}}
\caption{without lateral stepping reflex, the overturning and penetration}
\label{fig:22}
\end{figure}

We also compared robot's posture before, during and after the lateral stepping reflex to verify the effectiveness of the method. Fig.\ref{fig:21} shows the roll angle change of robot's body around y axis. The roll angle fluctuates between \(4.009^\circ\) and \(2.575^\circ\). Without the lateral stepping (called as reflex in the figure), Runner leans excessively and overturns under lateral impact. The roll angle in this condition changes as the red curve shows. Because we only establish the contact between the feet and the ground while setting up the virtual prototype, at \(t = 3.72s\), \(Roll = 90^\circ \), its torso touches the ground, penetrates and overturns until \(Roll = 180^\circ \) , seen in Fig.\ref{fig:22}. However, when the lateral stepping reflex is introduced, the roll angle changes like the green line in Fig.\ref{fig:21}. The partial enlarged drawing shows that although the roll angle suffers large fluctuation under lateral impact, it can be limited within a small range (\(-11.03^\circ  \sim 9.838^\circ \)) thanks to the lateral stepping reflex.

\begin{figure}[thpb]
\centering
  \includegraphics[width=84mm]{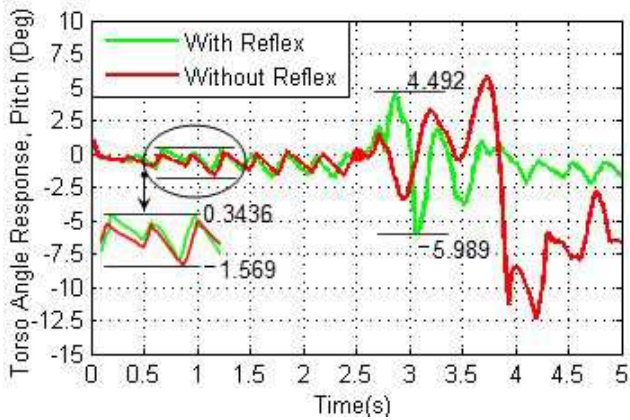}
\caption{the pitch angle change of robot's trunk}
\label{fig:23}       
\end{figure}

Similarly, the pitch angle change around x axis is shown in Fig.\ref{fig:23}. Normally, the pitch angle ranges from \(1.569^\circ\) to \(0.3436^\circ\). When the lateral impact is exerted, the fluctuation amplitude increases obviously. In Fig.\ref{fig:23}, the green line of pitch angle shows that the robot can remain stable with the help of lateral stepping reflex; the red line of pitch angle change indicates that the robot overturns without the reflex.

\begin{figure}[thpb]
\centering
\subfigure[Group1,from 50N to 300N]{
\includegraphics[width=3in]{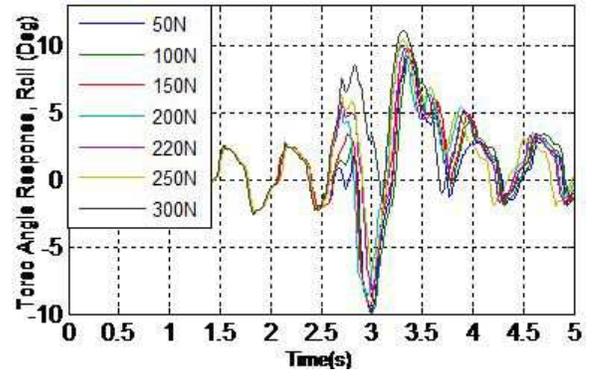}}
\subfigure[Group2,from 300N to 340N]{
\includegraphics[width=3in]{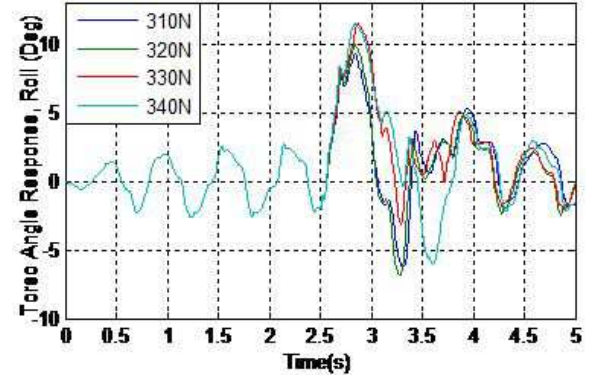}}
\caption{Different roll angle under varied lateral impact}
\label{fig:24}
\end{figure}

The simulation demonstrates that the robot can withstand impact no more than 160N without the reflex, but it can resist the maximum impact of 340N, 1.545 times of its weight with the reflex. The fact that it can withstand 2.125 times larger lateral impact than before attests to the effectiveness of our proposed method. In addition, to get robot's posture change rule under different impact, we exert impact of various magnitude on robot's side, and measured all the roll angle for comparison. The curves are shown in Fig.\ref{fig:24}.

We pick up 11 values from the scope of \(50N \sim 340N\). For the first group falling between 50N and 300N, the robot can return to normal within \(1\sim2\) cycles. As the impact increases, the changing range of roll angle also expands. When the robot takes the first lateral step, roll angle jumps to \( - 10^\circ \), indicating the occurrence of excessive lateral stepping, the degree will decline with the enlargement of impact magnitude. In the second group of \(310N \sim 340N\), Runner Robot will take over 2 gait cycle to return to normal. The roll angle reaches \(10^\circ \) and then \( - 7^\circ \), thus the degree of excessive lateral movement reduces to some extent compared to that of the first group.

\section{Conclusion and Future Work}
We put forward the CPG-based stability control method for quadruped robot under lateral impact: We equipped robot's four LM joints with oscillators with triggering enable property. With this extended CPG network, we introduced the lateral stepping, which can be triggered at large lateral acceleration. Based on the proper assignment of connecting weight matrix K, the output signal of CPG network can maintain the correct phase relation at any moment. The ZMP and inverted pendulum model are introduced to calculate the lateral step length. The simulation result confirmed that the robot can laterally move with trot gait with proper step length, and finally resume the stable state when receiving lateral impact. The comparison demonstrated that such method greatly improved robot's stability: it can withstand lateral impact of 340N instead of the previous 160N. Through a series of experiments of different magnitudes of lateral impact, we have found that the variation scope of roll angle enlarges as the lateral impact increases. In addition, small lateral impact will lead to excessive lateral movement, which we will study and discuss in our further experiment in the real world.




\section*{ACKNOWLEDGMENT}

This research was supported by the 2012 National Undergraduate Training Program for Innovation and Entrepreneurship.



\begin{thebibliography}{99}

\bibitem{c1} Sakagami Y, Watanabe R, Aoyama C, et al. The intelligent ASIMO: System overview and integration[C]//Intelligent Robots and Systems, 2002. IEEE/RSJ International Conference on. IEEE, 2002, 3: 2478-2483.
\bibitem{c2} Endo G, Nakanishi J, Morimoto J, et al. Experimental studies of a neural oscillator for biped locomotion with QRIO[C]//Robotics and Automation, 2005. ICRA 2005. Proceedings of the 2005 IEEE International Conference on. IEEE, 2005: 596-602.
\bibitem{c3} Kimura H, Fukuoka Y, Konaga K. Adaptive dynamic walking of a quadruped robot using a neural system model[J]. Advanced Robotics, 2001, 15(8): 859-878.
\bibitem{c4} Kimura H, Fukuoka Y. Biologically inspired dynamic walking of a quadruped robot on irregular terrain¡ªadaptation at spinal cord and brain stem[C]//Proceedings of the First International Symposium on Adaptive Motion of Animals and Machines. 2000.
\bibitem{c5} Miyakoshi S, Taga G, Kuniyoshi Y, et al. Three dimensional bipedal stepping motion using neural oscillators-towards humanoid motion in the real world[C]//Intelligent Robots and Systems, 1998. Proceedings., 1998 IEEE/RSJ International Conference on. IEEE, 1998, 1: 84-89.
\bibitem{c6} Taga G. A model of the neuro-musculo-skeletal system for anticipatory adjustment of human locomotion during obstacle avoidance[J]. Biological Cybernetics, 1998, 78(1): 9-17.
\bibitem{c7} Nagashima Nagashima F. A motion learning method using CPG/NP[C]//Proceedings of the 2nd International Symposium on Adaptive Motion of Animals and Machines. 2003: 4-8.
\bibitem{c8} Kimura and Fukuora   Kimura H, Fukuoka Y, Cohen A H. Adaptive dynamic walking of a quadruped robot on natural ground based on biological concepts[J]. The International Journal of Robotics Research, 2007, 26(5): 475-490.
\bibitem{c9} Raibert M, Blankespoor K, Nelson G, et al. Bigdog, the rough-terrain quadruped robot[C]//Proceedings of the 17th World Congress. 2008: 10823-10825.
\bibitem{c10} Kimura H., Witte H., Taga G. Briefing of AMAM[C]. Proc. Of Int. Symp. On Adaptive Motion of Animals and Machines. AMAM, Montreal, Canada. 2000.
\bibitem{c11} Ijspeert A J. Central pattern generators for locomotion control in animals and robots: a review[J]. Neural Networks, 2008, 21(4): 642-653.
\bibitem{c12} Kimura H., Fukuoka Y. Adaptive dynamic walking of a quadruped ¨C sensory feedback to CPG[C]. Neuromorphic Workshop, 2001.
\bibitem{c13} Cristina P.Santos, Vitor Matos, Gait transition and modulation in a quadruped robot: A brainstem-like modulation approach, Robotics and Autonomous Systems 59(2001)620-624.
\bibitem{c14} Vukobratovic M, Borovac B. Zero-moment point: Thirty five years of its life[J]. International Journal of Humanoid Robotics, 2004, 1(1)157-173.
\bibitem{c15} Barkan Ugurlu, Atsuo Kawamura. Bipedal Trajectory Generation Based on Combing Inertial Forces and Intrinsic Angular Momentum Rate Changes: Eulerian ZMP Resolution, IEEE TRANSACTIONS ON ROBOTICS. VOL.28.NO.6, December 2012.
\bibitem{c16} Wang G. Biped robot balance control¡ªBased on FRP feedback mechanism and ZMP[C]//Computer Science and Education (ICCSE), 2013 8th International Conference on. IEEE, 2013: 251-254.
\bibitem{c17} Liu J, Veloso M. Online ZMP sampling search for biped walking planning[C]//Intelligent Robots and Systems, 2008. IROS 2008. IEEE/RSJ International Conference on. IEEE, 2008: 185-190.
\bibitem{c18} Antonelli G. Stability analysis for prioritized closed-loop inverse kinematic algorithms for redundant robotic systems[J]. Robotics, IEEE Transactions on, 2009, 25(5): 985-994.

\end{thebibliography}
\end{document}